\documentclass{article}

\usepackage[final]{neurips_2025_ml4ps}

\usepackage[utf8]{inputenc} 
\usepackage[T1]{fontenc}    
\usepackage{hyperref}       
\usepackage{url}            
\usepackage{booktabs}       
\usepackage{amsfonts}       
\usepackage{gensymb}        
\usepackage{graphicx}       
\usepackage{nicefrac}       
\usepackage{microtype}      
\usepackage{xcolor}         
\usepackage{amsmath}

\usepackage{graphicx}
\usepackage{float}
\usepackage{soul}
\usepackage{xspace}
\usepackage{multirow}

\usepackage{todonotes}

\title{Diffusion-Based Electromagnetic Inverse Design\\of Scattering Structured Media}

\author{
  Mikhail Tsukerman\\
  Department of Physics\\
  ITMO University\\
  St. Petersburg 197101, Russian Federation\\
  \texttt{mik.tsukerman@gmail.com} \\
  \And
  Konstantin Grotov\\
  School of Electrical Engineering\\
  Tel Aviv University \\
  Tel Aviv 6997801, Israel \\
  \texttt{konstantin.grotov@gmail.com}\\
  \And
  Pavel Ginzburg \\
  School of Electrical Engineering\\
  Tel Aviv University \\
  Tel Aviv 6997801, Israel \\
  \texttt{pginzburg@tauex.tau.ac.il}
}

\begin{document}

\maketitle

\begin{abstract}
We present a conditional diffusion model for electromagnetic inverse design that generates structured media geometries directly from target differential scattering cross-section profiles, bypassing expensive iterative optimization. Our 1D U-Net architecture with Feature-wise Linear Modulation learns to map desired angular scattering patterns to $2\times2$ dielectric sphere structure, naturally handling the non-uniqueness of inverse problems by sampling diverse valid designs. Trained on 11,000 simulated metasurfaces, the model achieves median MPE below 19\% on unseen targets (best: 1.39\%), outperforming CMA-ES evolutionary optimization while reducing design time from hours to seconds. These results demonstrate that employing diffusion models is promising for advancing electromagnetic inverse design research, potentially enabling rapid exploration of complex metasurface architectures and accelerating the development of next-generation photonic and wireless communication systems. The code is
publicly available at \url{https://github.com/mikzuker/inverse_design_metasurface_generation}.
\end{abstract}

\section{Introduction}

Engineered metasurfaces enable precise control of electromagnetic waves, with applications spanning high-resolution imaging, compact optical devices~\citep{Hu2024Jun, Faraji-Dana2018Oct}, and next-generation wireless communications~\citep{Hajiaghajani2021Nov}. However, inverse design---finding a structure that produces an object scattering response---is challenging due to nonlinear boundary conditions, high dimensionality, and the one-to-many nature of the problem~\citep{li_raphaël_pestourie_lin_johnson_capasso_2022}. Traditional methods, such as topology optimization or genetic algorithms, rely on iterative simulations, often requiring expert tuning while incurring prohibitive computational costs~\citep{Qian2020Sep}. Furthermore, the complex design space complicates brute-force search, making data-driven approaches appealing~\citep{Lee2023Jul}.

Machine learning has emerged as a promising alternative, learning the structure–response mapping from data to bypass iterative solvers~\citep{Bastek2022Jan}. Generative models, in particular, address the one-to-many challenge by sampling from the distribution of viable designs rather than predicting a single solution~\citep{Wang2020Jun}. For instance, VAEs and GANs have successfully generated metamaterials with tailored optical or mechanical properties~\citep{Pahlavani2022Dec}, and diffusion models, recently dominant in high-dimensional generation, offer stable training and diverse outputs~\citep{Ho2020Jun}. Early applications in photonics and mechanics suggest that diffusion models can capture complex physics, enabling fast, accurate inverse design~\citep{An2019Aug}.

Building on these successes across physical domains, we investigate whether diffusion models can similarly be applied to electromagnetic inverse design. We propose a diffusion-based framework for inverse scattering design, focusing on conditional generation of metasurfaces from target differential scattering cross-section (DSCS) profiles. In this work, we specifically generate structured media consisting of dielectric spheres arranged in $2\times2$ grids. Our one-dimensional diffusion model maps desired angular scattering patterns to metasurface structures, explicitly handling nonuniqueness by producing diverse valid designs. The trained model generalizes to novel out-of-distribution targets and exceeds the performance of the evolutionary optimization algorithm.

\section{Problem}

Natural and engineered objects exhibit complex electromagnetic scattering patterns that are challenging to reproduce directly. In the electromagnetic community, metasurfaces---man-made engineered structures, are considered as a solution in cases where electromagnetic wave propagation shall be controlled at will. However, the inverse problem of finding a structure tailoring a specific electromagnetic response remains a challenge and, in many cases, leads to over-complicated designs. This motivates our inverse design problem: Given a set of differential scattering cross-section (DSCS) values at predefined polar angles, generate a structured medium geometry that reproduces the target scattering profile. While our approach specifically generates structured media composed of dielectric spheres, we adopt the term ``metasurface'' throughout this work for consistency with established terminology in the electromagnetic community.

\begin{figure}[h]
    \centering
    \includegraphics[width=\linewidth]{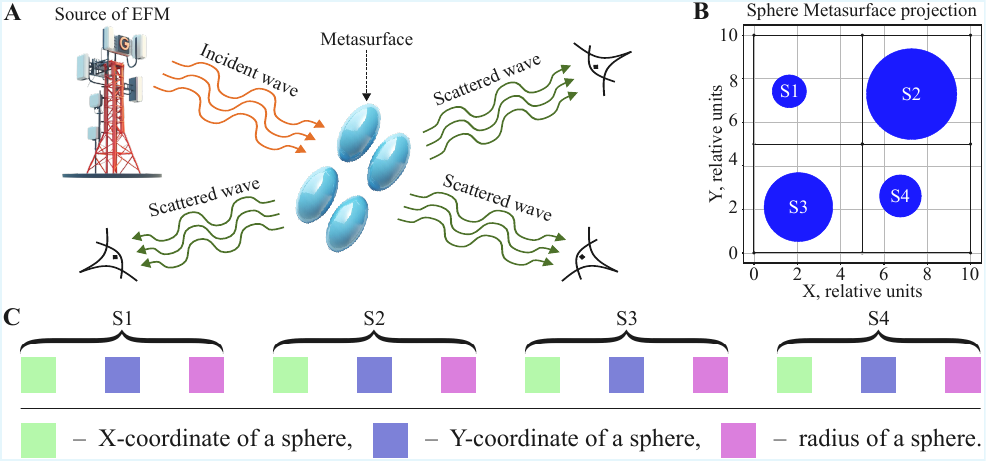}
    \caption{(A) Schematic outline of the problem; (B) Parametrization of metasurface with spheres placed in a $2 \times 2$ grid; (C) Schematic of the encoded geometry vector for a metasurface.}
    \label{fig:metasurface_example}
\end{figure}

In this work, we focus on a specific class of metasurfaces composed of dielectric spheres arranged on a regular grid. The parametrization of our metasurface geometry is illustrated in Figure~\ref{fig:metasurface_example}(B). A square virtual substrate is discretized into an $N \times N$ grid of square cells (black lines). Within each cell, we place a single dielectric sphere with a fixed refractive index (treated as a hyperparameter). Each sphere is characterized by three parameters: its center position $(x, y)$ and radius $r$, all defined relative to the enclosing cell boundaries. For simplicity, in this study, we focus on the metasurfaces with $N=2$, and leave larger grids for future research.

For further training, we encode these geometric parameters into a one-dimensional representation suitable for neural network processing. As illustrated in Figure~\ref{fig:metasurface_example}(C), each metasurface is represented as a vector of dimension $3N^2$, where consecutive triplets $(p_x, p_y, p_r)$ encode the relative horizontal position, vertical position, and radius of each sphere. All parameters are normalized to the range $[0, 1]$, ensuring consistent scaling across different metasurface configurations and stabilizing gradient-based optimization.
This normalized representation enables it to learn patterns across diverse geometries while ensuring that the generated structures are physically valid.

The complete illustration of the problem components is presented in Figure~\ref{fig:metasurface_example}(A).

\section{Diffusion Model for Inverse Design}

In this section, we present our conditional diffusion-based approach for inverse metasurface design. We first describe the forward electromagnetic solver and dataset generation process, then detail our model architecture and training methodology.

\paragraph{Forward solver and dataset generation} The foundation of any data-driven inverse design approach is a reliable forward solver that calculates electromagnetic responses based on geometric parameters. We employ the SMUTHI package~\citep{egel2021smuthi}, which efficiently computes electromagnetic scattering from spherical objects using T-matrices~\citep{schulz_stamnes_stamnes_1998}. This solver provides fast and accurate computation of differential scattering cross-sections for our metasurface configurations, enabling large-scale dataset generation. Using this framework, we generated a dataset of metasurfaces $2 \times 2$, consisting of 11,000 unique samples with their corresponding DSCS values calculated at 10 polar angles.

\begin{figure}[h!]
    \centering
    \includegraphics[width=\linewidth]{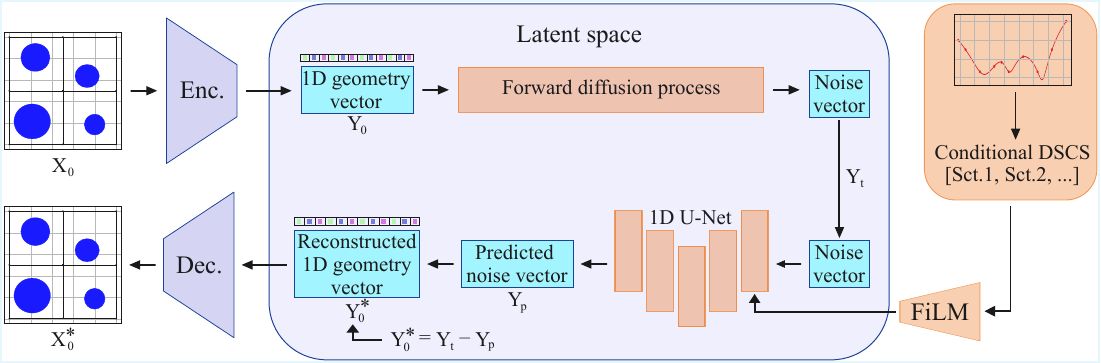}
    \caption{Architecture of the conditional diffusion model for metasurface generation.}
    \label{fig:architecture}
\end{figure}

\paragraph{Model architecture} To address the one-to-many nature of inverse scattering, we employ a conditional diffusion model that learns to generate metasurface geometries from target DSCS profiles. As illustrated in Figure~\ref{fig:architecture}, the model operates through two complementary phases: a forward process that gradually adds noise to training examples, and a learned reverse process that iteratively denoises initial random configurations while respecting the conditioning DSCS values. 

The denoising is performed by a 1D U-Net~\citep{ronneberger2015u} architecture that predicts noise removal operations at each step, with target scattering profiles incorporated through Feature-wise Linear Modulation (FiLM)~\citep{Perez2017Sep} of the network's layers. This conditioning mechanism allows the model to adapt its generation process to specific electromagnetic response requirements. The complete mathematical formulation of the diffusion process is provided in Appendix~\ref{app:model_details}.

\paragraph{Training and evaluation} We train the model using the standard diffusion loss that minimizes the expected $L_2$ distance between predicted and true noise across all timesteps, with training hyperparameters detailed in Appendix~\ref{appendix:parameters}. To quantitatively assess generation quality, we evaluate the Mean Percentage Error (MPE) between conditioning input and generated structure DSCS values.

For a more detailed analysis of the MPE evolution during training, we saved intermediate versions of the model every $1000$ steps. Using these checkpoints, we sampled $10$ metasurfaces for the inverse design of a randomly generated spectrum. 

\section{Experimental Results and Discussion}

\begin{figure}[h!]
    \centering
    \includegraphics[width=\linewidth]{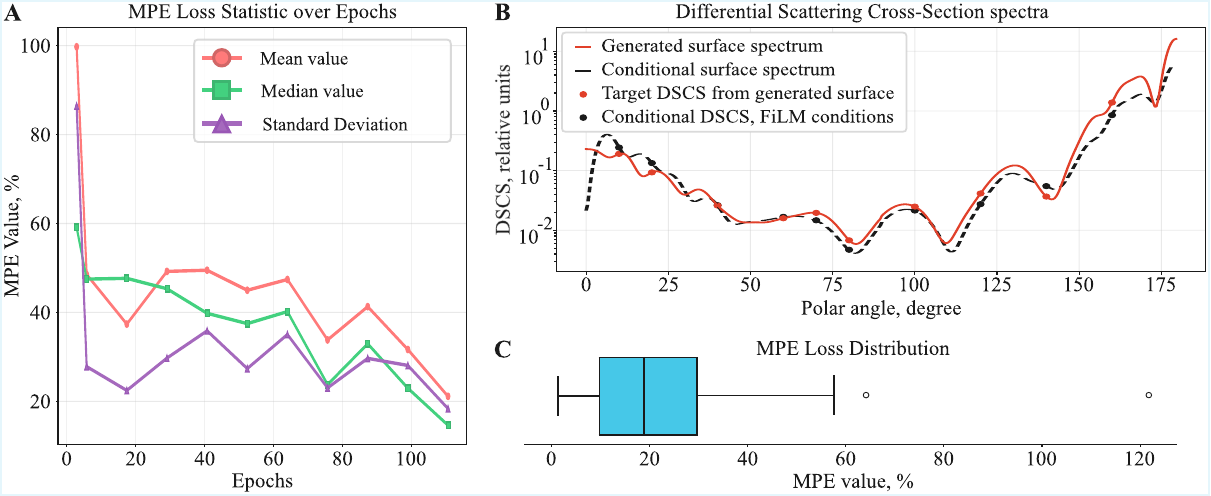}
    \caption{Performance of the trained diffusion model. (A) MPE statistics over training epochs. (B) DCSC spectra comparison between ground-truth (black dashed line) and best generated metasurface sample with MPE = 1.39\% (red solid line). (C) MPE distribution for 40 generated samples for selected out-of-distribution conditioning.}
    \label{fig:results}
\end{figure}

To evaluate our diffusion model's performance, we tracked the MPE evolution throughout training and tested its ability to generate novel metasurfaces. As shown in Figure~\ref{fig:results}(A), all three statistical measures—mean, median, and standard deviation—exhibit consistent decline over 116 epochs. This convergence pattern indicates that the model successfully learns the physics-to-geometry mapping inherent in electromagnetic scattering.

To test generalization beyond the training set, we evaluated the model on a metasurface configuration not seen during training. This experiment assesses whether the model has learned the underlying electromagnetic principles rather than memorizing training examples. We sampled 40 candidate metasurfaces conditioned on the target DSCS profile of this unseen structure. The best-performing sample (Figure~\ref{fig:results}(B)) achieved an MPE of 1.39\%, demonstrating accurate reproduction of the target scattering response at the specified angles.

The distribution of MPE values across all 40 generated samples (Figure~\ref{fig:results}(C)) provides insights into model robustness. With a median error of 18.91\% and a relatively tight interquartile range, the model shows consistent performance without significant outliers. While the generated metasurfaces accurately match the target DSCS at the 10 specified conditioning angles, they naturally differ from the ground-truth structure at other angles---a consequence of the fundamental non-uniqueness in inverse scattering problems (see Appendix~\ref{appendix:geometries_comparison} for geometric comparison).

One of the major advantages of using the diffusion model for electromagnetic inverse design is its fast inference speed compared to generating designs through optimization processes. However, this comparison would not be fair without taking into account the training time of the diffusion model. We compare our model's training time and performance with the commonly used evolutionary optimization algorithm. For this, we perform inverse design optimization using the CMA-ES~\citep{hansen1996adapting} algorithm, which is widely employed in various electromagnetic optimization problems~\citep{martinez2013ultra, gregory2011fast}. We use the Python implementation from~\citep{nomura2024cmaes} with parameters detailed in Appendix~\ref{appendix:cmaes_parameters}, which were derived from successful experiments in recent studies.

We observe that in addition to better evaluation results (3\% MPE for the diffusion model vs. 5\% for CMA-ES), the diffusion model offers significant computational savings (6 hours for one-time training followed by seconds per generation, compared to 15-20 hours per optimization for CMA-ES across four seeds). This significant gap stems from the objective function calculation (using a forward solver) required during the evolutionary optimization algorithm, which is not needed during the diffusion model's training.

However, this comparison requires careful interpretation. For CMA-ES, the typical computational cost scales with problem dimensionality: the recommended population size is $\lambda = 4 + \lfloor 3\ln(d) \rfloor$ \citep{Hansen2016Apr}, and convergence typically requires $O(100d)$ to $O(1000d)$ iterations \citep{Hansen2003}, yielding approximately $10^4$ to $10^5$ forward solver evaluations per run. In our experiments, this resulted in approximately $1.05 \times 10^5$ evaluations per optimization run (1,500 iterations $\times$ 70 population size). In contrast, the diffusion model requires only $1.1 \times 10^4$ evaluations to generate its training dataset (a one-time cost). Once trained, the diffusion model can generate new designs without additional solver calls, whereas CMA-ES must repeat the full optimization for each new problem instance. Thus, the diffusion approach amortizes its computational cost across multiple inference tasks, making it increasingly advantageous when solving multiple related inverse problems.

While this is still not a complete comparison of two fundamentally different methodologies, we consider it a promising direction for electromagnetic inverse design.

Although demonstrated here for $2\times2$ structured media, our approach readily scales to higher-dimensional designs, where the computational advantages over iterative optimization would be even more significant. These results establish diffusion models as a promising direction for accelerating electromagnetic inverse design workflows.

\bibliographystyle{plainnat}
\bibliography{main_text} 

\appendix

\section{Model Implementation Details}
\label{app:model_details}

\paragraph{Diffusion Process} Our model follows the DDPM framework \citep{Nakkiran2024Jun}, with:
\begin{equation*}
\label{equation:forward}
y_t = \sqrt{\bar{\alpha}_t} \cdot y_0 + \sqrt{1 - \bar{\alpha}_t} \cdot \epsilon
\end{equation*}
using a cosine noise schedule \citep{Nichol2021Feb}:
\begin{equation*}
f(\tau) = \cos^2(\dfrac{\tau + s}{1+s}\cdot\dfrac{\pi}{2}); \text{ } \bar{\alpha}_t = \dfrac{f(\frac{t}{T})}{f(0)}
\end{equation*}

\paragraph{Reverse Process} The denoising model learns ($\varepsilon_\theta(y_t, t)$ is predicted by the U-Net noise):
\begin{equation*}
\hat{y}_0 \;=\; \frac{y_t - \sqrt{1-\bar{\alpha}_t} \cdot \varepsilon_\theta(y_t, t)}{\sqrt{\bar{\alpha}_t}} 
\end{equation*}

Using $\hat{y}_0$ the model is then denoise step $t-1$:

\begin{equation*}
\mu_\theta(y_t,t) = \frac{\sqrt{\bar\alpha_{t-1}}\,\beta_t}{1-\bar\alpha_t}\,\hat{y}_0
+ \frac{\sqrt{\alpha_t}\,(1-\bar\alpha_{t-1})}{1-\bar\alpha_t}y_t 
\end{equation*}
\begin{equation*}
y_{t-1} = \mu_\theta(y_t,t) + \sqrt{\frac{1-\bar\alpha_{t-1}}{1-\bar\alpha_t}\,\beta_t} \cdot z,
\text{ } z \sim \mathcal{N}(0,I)
\end{equation*}

where parameters follow:
\begin{equation*}
\alpha_t = 1 - \beta_t, \text{ }
\beta_t = 1 - \frac{\bar{\alpha}_t}{\bar{\alpha}_{t-1}}
\end{equation*}

\paragraph{FiLM Conditioning} The U-Net uses feature-wise transformations:
\begin{equation*}
\text{FiLM}(F_{i,c}) = \gamma_{i,c} \cdot F_{i,c} + \beta_{i,c}
\end{equation*}
with $\gamma_{i,c}, \beta_{i,c}$ generated by two-layer networks processing the DSCS conditions.

\section{Model Training Parameters}
\label{appendix:parameters}
The model was trained using the parameters shown in Table \ref{table:parameters}.
 
\begin{table}[h!]
\centering
\caption{Hyperparameters of the trained diffusion model.}
\begin{tabular}{llc}
\hline
\textbf{Category} & \textbf{Parameter} & \textbf{Value} \\ \hline
\multirow{3}{*}{Architecture} 
    & U-Net dimensions & \{16, 32, 64, 128, 128, 64, 32, 16\} \\
    & Conditional vector size & 10 \\
    & FiLM hidden dimension & 128 \\ \hline
\multirow{2}{*}{Diffusion Process} 
    & Number of denoising steps & 1000 \\
    & EMA decay coefficient & 0.995 \\ \hline
\multirow{3}{*}{Training} 
    & Learning rate & $4 \times 10^{-6}$ \\
    & Batch size & 16 \\
    & Epochs & 116 \\ \hline
\end{tabular}
\label{table:parameters}
\end{table}

\section{Ground Truth vs Generated Metasurface Comparison}
\label{appendix:geometries_comparison}

\begin{figure}[h!]
    \centering
    \includegraphics[width=0.7\linewidth]{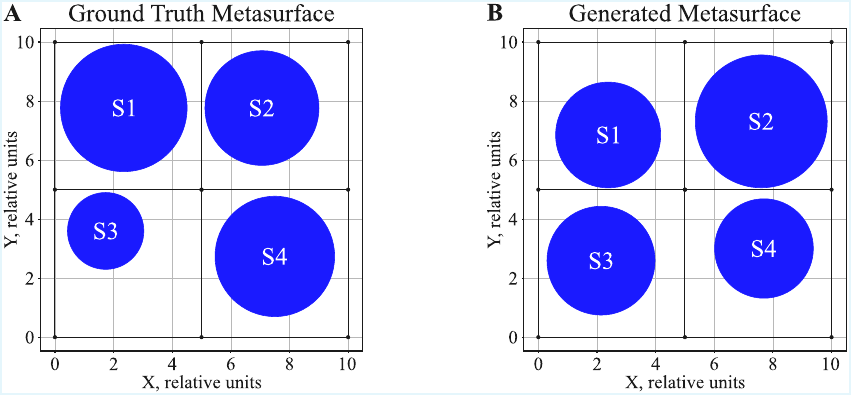}
    \caption{Comparison of the ground truth and generated metasurface.}
    \label{fig:comparison}
\end{figure}

To illustrate the non-uniqueness inherent in inverse scattering problems, Figure~\ref{fig:comparison} compares the ground truth metasurface with our best generated sample. Figure~\ref{fig:comparison}(A) shows the original structure from which target DSCS values were extracted at ten specified polar angles. Figure~\ref{fig:comparison}(B) presents the generated metasurface achieving 1.39\% MPE on these angles. Despite matching scattering responses, the structures exhibit distinct geometries---demonstrating that multiple physical realizations can produce nearly identical electromagnetic signatures at selected observation angles.

\section{Feasibility of Generated Structures}
\label{appendix:feasibility}

Although all simulation parameters are expressed relative to the wavelength of the incident wave, they can be readily converted into physical units once the wavelength is specified. For instance, in our experiments, the wavelength was normalized to 1, and the unit cell length of the generated metasurfaces was set to 5. The refractive index of the spheres composing the metasurface was equal to 2. By defining one relative unit as 3 cm, we find that the incident wave's frequency corresponds to approximately 10 GHz, and the unit cell length corresponds to 15 cm (with the total metasurface side length equal to 30 cm). These frequency and size ranges fall within the radio frequency domain and can be feasibly realized under laboratory conditions, consistent with experimentally demonstrated metasurfaces of comparable dimensions (10 cm)~\citep{mikhailovskaya2025superradiant}.

\section{CMA-ES Parameters Used For Comparison}
\label{appendix:cmaes_parameters}

For comparison with our diffusion model, we employed the CMA-ES algorithm with parameters selected based on successful configurations from recent electromagnetic optimization studies~\citep{dobrykh20253d, mikhailovskaya2025superradiant}. The algorithm was initialized with a uniformly random mean vector in the normalized parameter space $[0,1]^{12}$, corresponding to the 12-dimensional geometry representation of our $2\times2$ structured media. The optimization was run for four independent seeds to account for the stochastic nature of the evolutionary search with the parameters shown in Table~\ref{table:cmaes_parameters}.

\begin{table}[h!]
\centering
\caption{CMA-ES algorithm parameters used for inverse design comparison.}
\begin{tabular}{cc}
\hline
\textbf{Parameter} & \textbf{Value} \\ \hline
Initial mean vector & Uniform random in $[0,1]^{12}$ \\
$\sigma$ & 0.07 \\
Population size & 70 \\
Iterations & 1500 \\ \hline
\end{tabular}
\label{table:cmaes_parameters}
\end{table}

\end{document}